\pdfoutput=1

\documentclass[11pt]{article}

\usepackage{hyperref}
\usepackage[]{ACL2023}

\usepackage{times}
\usepackage{latexsym}

\usepackage[T1]{fontenc}

\usepackage[utf8]{inputenc}

\usepackage{microtype}

\usepackage{inconsolata}

\usepackage{subfigure}

\usepackage{booktabs} 
\usepackage{amsmath}
\usepackage{xspace}
\usepackage{amssymb}
\usepackage{graphicx}
\usepackage{multirow}
\usepackage{enumitem}
\usepackage{ulem}
\usepackage{float}
\usepackage{makecell}

\DeclareMathOperator{\softmax}{softmax}

\newcommand{\ie}{\textit{i.e.,}\xspace}
\newcommand{\eg}{\textit{e.g.,}\xspace}

\newcommand{\modelname}{FreeLM\xspace}

\newcommand{\datasetnum}{30\xspace}

\title{FreeLM: Fine-Tuning-Free Language Model}

\author{
  Xiang Li\textsuperscript{1\textdagger}, Xin Jiang\textsuperscript{1\textdagger}, Xuying Meng\textsuperscript{2}, Aixin Sun\textsuperscript{3}, Yequan Wang\textsuperscript{1$*$}\\
  $^{1}$Beijing Academy of Artificial Intelligence, Beijing, China\\
  $^{2}$Institute of Computing Technology, Chinese Academy of Sciences, Beijing, China\\
  $^{3}$School of Computer Science and Engineering, Nanyang Technological University, Singapore\\
  \texttt{lixiang@baai.ac.cn, jiangxin@baai.ac.cn, mengxuying@ict.ac.cn} \\
  \texttt{axsun@ntu.edu.sg, tshwangyequan@gmail.com}
}

\begin{document}
\maketitle

\renewcommand{\thefootnote}{\fnsymbol{footnote}}
\footnotetext[1]{Corresponding author}
\footnotetext[2]{Indicates equal contribution}
\renewcommand{\thefootnote}{\arabic{footnote}}

\begin{abstract} 
Pre-trained language models~(PLMs) have achieved remarkable success in NLP tasks.
Despite the great success, mainstream solutions largely follow the \textit{pre-training then fine-tuning} paradigm, which brings in both high deployment costs and low training efficiency. 
Nevertheless, fine-tuning on a specific task is essential because PLMs are only pre-trained with \textit{language signal} from large raw data.  
In this paper, we propose a novel \textit{fine-tuning-free} strategy for language models, to consider both \textit{language signal} and \textit{teacher signal}. \textit{Teacher signal} is an abstraction of a battery of downstream tasks, provided in a unified proposition format. Trained with both language and strong task-aware teacher signals in an interactive manner, our 
\modelname model demonstrates strong generalization and robustness. 
\modelname outperforms large models \eg GPT-3 and InstructGPT, on a range of language understanding tasks in experiments.  \modelname is much smaller with 0.3B parameters, compared to 175B in these models.   

\end{abstract}

\section{Introduction}
\label{sec:intro}

Pre-trained language models (PLMs), exampled by BERT~\cite{devlin-etal-2019-bert}, the GPT series~\cite{radford2018improving, radford2019language,brown2020language} and their variants~\cite{DBLP:journals/corr/abs-1907-11692,DBLP:journals/tacl/JoshiCLWZL20,DBLP:journals/corr/abs-1904-09223,DBLP:journals/jmlr/RaffelSRLNMZLL20}, have been widely applied in various language processing tasks and achieved remarkable success.

\begin{figure}
    \centering
    \includegraphics[scale=0.44]{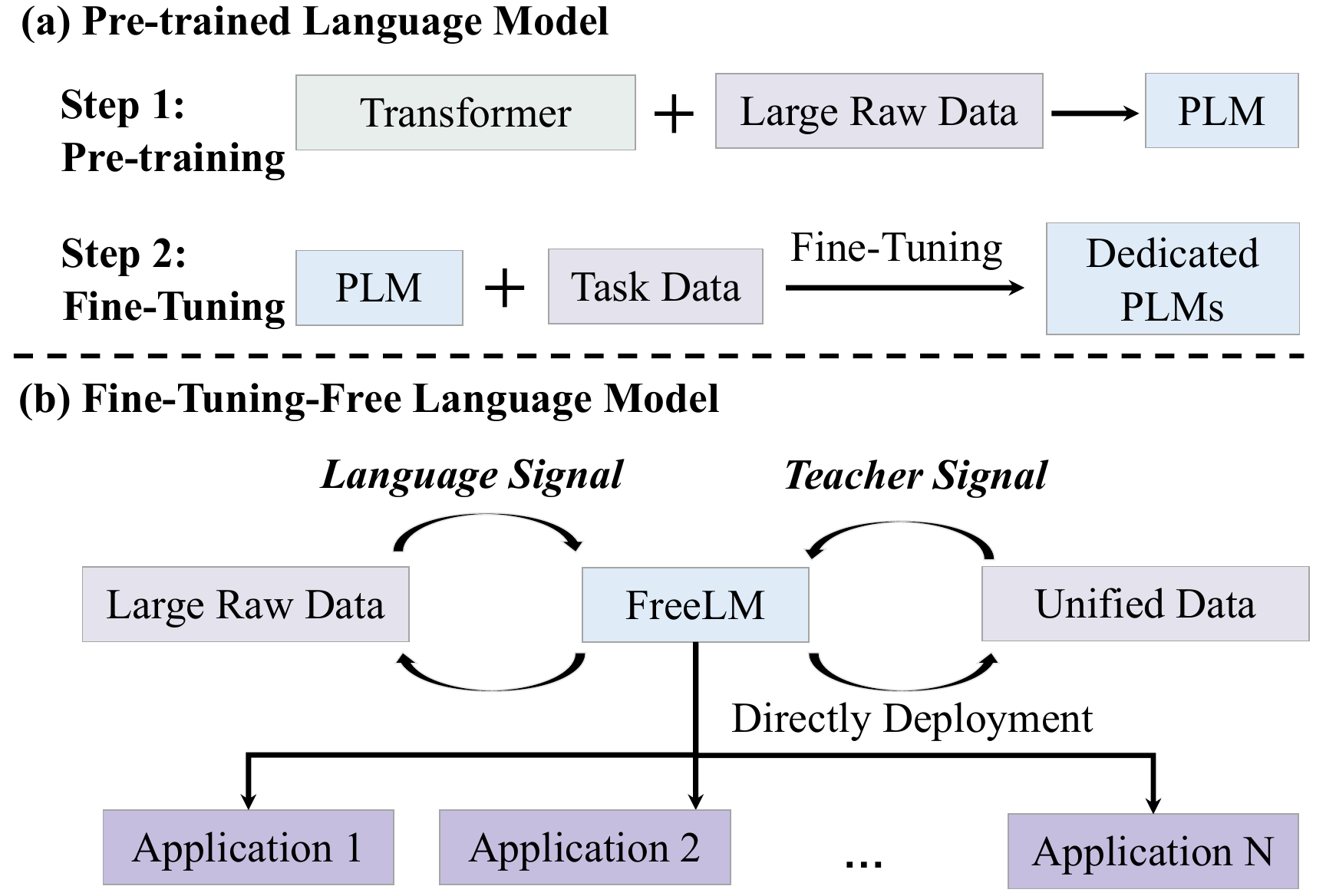}
    \caption{An overview of small models, pre-trained language models, and our proposed fine-tuning-free language model. }
    \label{fig:intro}
    \vskip -1em
\end{figure}

Despite their great success, the \textit{pre-training then fine-tuning}  paradigm \cite{devlin-etal-2019-bert,radford2019language} brings in very large costs for training and deployment in enterprise-level applications. Even large companies are very careful in using billion-parameter PLMs online~\cite{distilbert}, and remain showing high interest in small models. 
Before the era of PLMs, small models are trained for specific tasks. For some tasks, these task-dedicated models may perform comparably to or even better than large PLMs with fine-tuning~\cite{DBLP:journals/corr/abs-2207-08815}. Both the training and deployment costs are manageable since the task-specific datasets and models are typically much smaller. Nevertheless, additional efforts are essential to design task-specific models, one for each task. Illustrated in Figure~\ref{fig:intro}, pre-trained with much larger datasets, PLMs gain high generalization, and can be fine-tuned to diverse downstream tasks for language understanding and generation. The effort of designing task-specific small models is now replaced by the huge costs of pre-training on large datasets and the high deployment costs.

To reduce deployment costs, zero-shot \eg GPT-3~\cite{brown2020language} and few-shot models have been investigated. Their performance, particularly on understanding tasks, remains unsatisfactory. One of the reasons is that the PLMs are trained with only \textit{language signal} which is not task-aware, \ie the training objective of PLMs does not well align with the task objective.  
Recently, instruction-tuning-based models, \ie InstructGPT \cite{https://doi.org/10.48550/arxiv.2203.02155} and FLAN \cite{DBLP:conf/iclr/WeiBZGYLDDL22}, further improves zero-shot performance. Their main design transfers everything into language text, then uses self-supervised \textit{language signal} to train the model. In this way, the \textit{language signal} in their training becomes more task-aware and relatively stronger. However, to achieve the best performance on specific tasks, adaptation remains necessary.

The idea of enhancing task awareness in training motivates us to consider a \textit{reasonably sized model} that is capable of generalizing to \textit{a good range of pre-defined tasks}. The ultimate goal is to achieve good performance on all these pre-defined tasks (and potentially  unseen tasks) with low training and deployment costs. We believe this is an ideal setting for an enterprise to handle all their own specific tasks with a single model trained on their own data, without additional fine-tuning.

The key challenge then becomes how to unify the pre-defined tasks as a strong task-aware supervision signal for model training. To this end, we propose \modelname, a fine-tuning-free language model. Illustrated in Figure~\ref{fig:intro},  \modelname is trained with  both \textit{language signal} and \textit{teacher signal} in an iterative way. During training, in different iterations, the model alternately learns  (i) the \textit{language signal} from large raw data, and (ii) the \textit{teacher signal} from unified data prepared based on all pre-defined tasks. 
Benefiting from the interplay between both signals,  \modelname is able to handle all pre-defined tasks with high accuracy, and handle unseen tasks well as a typical pre-trained language model does.

In addition to the \textit{fine-tuning-free} strategy, a major contribution is that we make the very first attempt to unify a good range of language tasks through a unified data format, called \textit{proposition format}. Datasets constructed for all pre-defined tasks are then mapped to this unified proposition format, and become the unified data in Figure~\ref{fig:intro}. Through the proposition format, we provide proposition correctness judgment to help the language model know the facts. The unified format also implicitly facilitates mutual improvements among these tasks, which further contributes to better generalization and robustness.

To the best of our knowledge, this is the first attempt to propose an effective \textit{fine-tuning-free} strategy for large language model training. We also demonstrate how to unify multiple tasks through a unified format to make the PLM more task-aware through strong supervision signals. Through experiments, we show that \modelname greatly outperforms state-of-the-art models  on language understanding tasks, compared to GPT-2, 175B GPT-3 and InstructGPT. We also show that the generation ability of \modelname is even better than GPT-2. \modelname achieves good performance on unseen tasks, and has excellent generalization and robustness.


\section{Related Work}
\label{sec:related}

In this work, we focus on both language generation and language understanding tasks. Thus, we only consider unidirectional Transformers like GPT as basic models, and do not consider masked language models (\eg BERT) or encoder-decoder language models (\eg T5). Next, we brief auto-regressive language models and instruct-tuning-based models.

\subsection{Auto-Regressive Language Models}

Auto-regressive language models are trained to predict the next token based on all previous tokens. The unidirectional nature empowers these models with language generation capability. 
The impacts of model scale and model structure are the two key research directions. 

On model scale, the most typical and also the most influential auto-regressive LMs are the GPT series, \ie GPT-1~\cite{radford2018improving}, GPT-2~\cite{radford2019language}, and GPT-3~\cite{brown2020language}.
In particular, the success of GPT-3 has made researchers realize that the violent aesthetics of the model scale and large raw data can have such a good generation performance. 
To make the models bigger, even methods like MoE~\cite{MOE2} are proposed. However, large scale brings in large costs and many challenges, particularly the high costs of fine-tuning on downstream tasks.

To improve the model structure, GLM~\cite{GLM} is designed to utilize the autoregressive blank infilling. 
Transformer-XL~\cite{DBLP:conf/acl/DaiYYCLS19} tries to solve the longer-term dependency, and its variant XL-Net~\cite{DBLP:conf/nips/YangDYCSL19} hopes to learn bidirectional contexts. 
However, these models do not perform better than GPT-3 as the models get larger. The vital reason remains how to effectively and efficiently utilize data.

\subsection{Instruct-Tuning Based Models}

A larger model size does not mean that it can produce output that better meets user expectations~\cite{https://doi.org/10.48550/arxiv.2203.02155}. 
One solution is to fine-tune large language models based on a wide range of tasks, human feedback, and so on. The typical and most representative model is InstructGPT~\cite{https://doi.org/10.48550/arxiv.2203.02155}. Through reinforcement learning, data from tasks, and human feedback, InstructGPT achieves excellent performance on both language understanding and generation tasks. Experiments even show that InstructGPT can align with humans compared with original GPT-3. EFL~\cite{https://doi.org/10.48550/arxiv.2104.14690}, designed for few-shot learning,  reformulates potential NLP tasks into an entailment task based on RoBERTa-large. Entailment task is hard to handle well more types of tasks, while our unified proposition is more generic to  fit more tasks. In addition, EFL suffers from degradation in generation abilities.

In our model design, we aim to achieve better results on both understanding and generation tasks without fine-tuning.
InstructGPT adopts a method like ``all in language''. EFL gives up the generation ability. To our understanding, these models do not effectively utilize the \textit{teacher supervision signal}. 
In this paper, we design an iterative training strategy on both language raw data and task-aware data. The key idea here is to teach the model to be task-aware data, while not forgetting its role of a language model, \ie to model language.

\section{Task Unification}
\label{sec:unified}

Our goal is to train a task-aware language model which learns from language as a typical PLM does, and also learns from a good number of task-specific datasets. Shown in Figure~\ref{fig:intro}, \modelname learns from both language data and unified data, where the latter is the result of unifying pre-defined tasks.

\paragraph{Language Data.} For language data, the choice is relatively straightforward. We adopt OpenWebText \cite{Gokaslan2019OpenWeb}, an open-source replication of the WebText~\cite{radford2019language} dataset proposed by OpenAI. It is built by scraping all outbound links from the social media platform Reddit, which received at least $3$ karmas. We use the same settings as GPT-2 on this dataset.

\paragraph{Unified Data in Proposition Format.} Our key motivation is to enable enterprises to handle all their own specific tasks with a single model trained on their own data, without additional fine-tuning. However, it is challenging to access and conduct experiments on proprietary data. Even if that is feasible, our results will be hard to be benchmarked against the current state-of-the-art. Without  loss of generality, we choose to unify seven well-defined and popular language tasks: 1. \textit{question answering}, 2. \textit{paraphrasing}, 
 3. \textit{topic classification}, 4. \textit{story cloze}, 5. \textit{sentiment classification}, 6. \textit{natural language inference}, and 7. \textit{linguistic acceptability}. 

We unify these seven tasks by transforming them into a ``proposition correctness judgment'' task, to judge whether a proposition is true. Specifically, for each task, we design some candidate templates, then use the designed rule to transfer each instance into a \textit{\textbf{proposition format}}: 

\begin{table}[h]
\centering 
\begin{tabular}{p{2.6in}}
\toprule
``[tsk] \textit{task name} [tsk] \textit{instance text(s)} [sep] \textit{dynamic prompt} [cls]''\\
\bottomrule
\end{tabular}
\end{table}

\noindent The \textit{task name} follows the original pre-define task name before unification like ``question answering''. Because each task has its own characteristic, hence we use \textit{dynamic prompt} to fit them well considering the ground-truth label.

\begin{figure}[t]
    \includegraphics[scale=0.65]{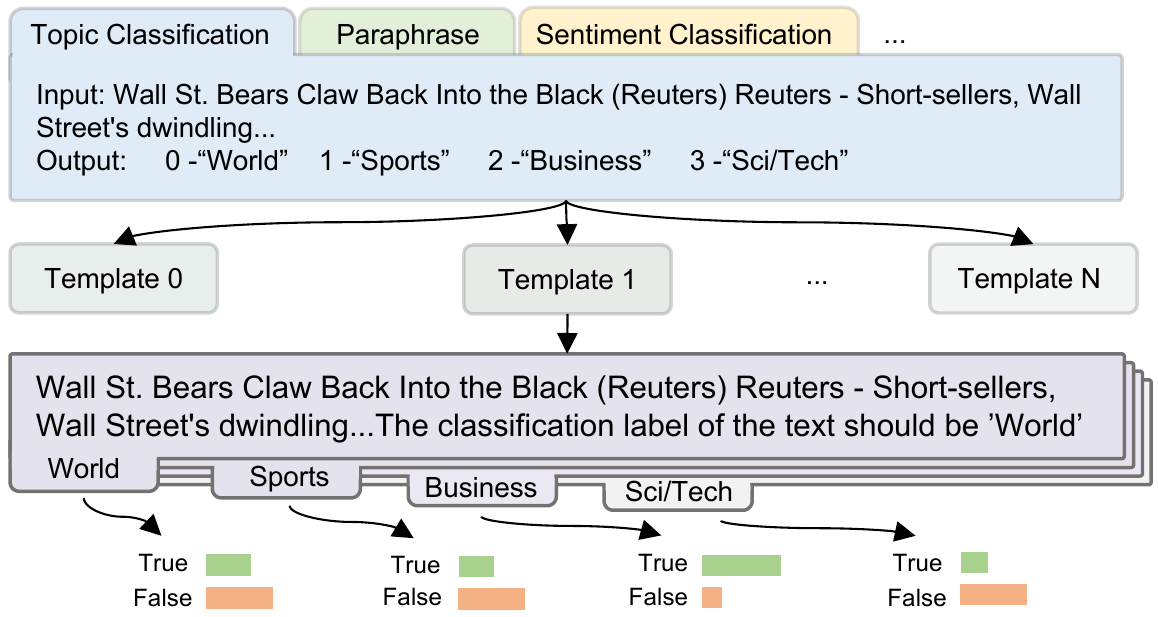}
    \caption{A template example for the topic classification task. It also describes the transform rule for task unification.}
    \label{fig:template}
\end{figure}

\begin{figure}[t]
    \centering
    \includegraphics[scale=0.44]{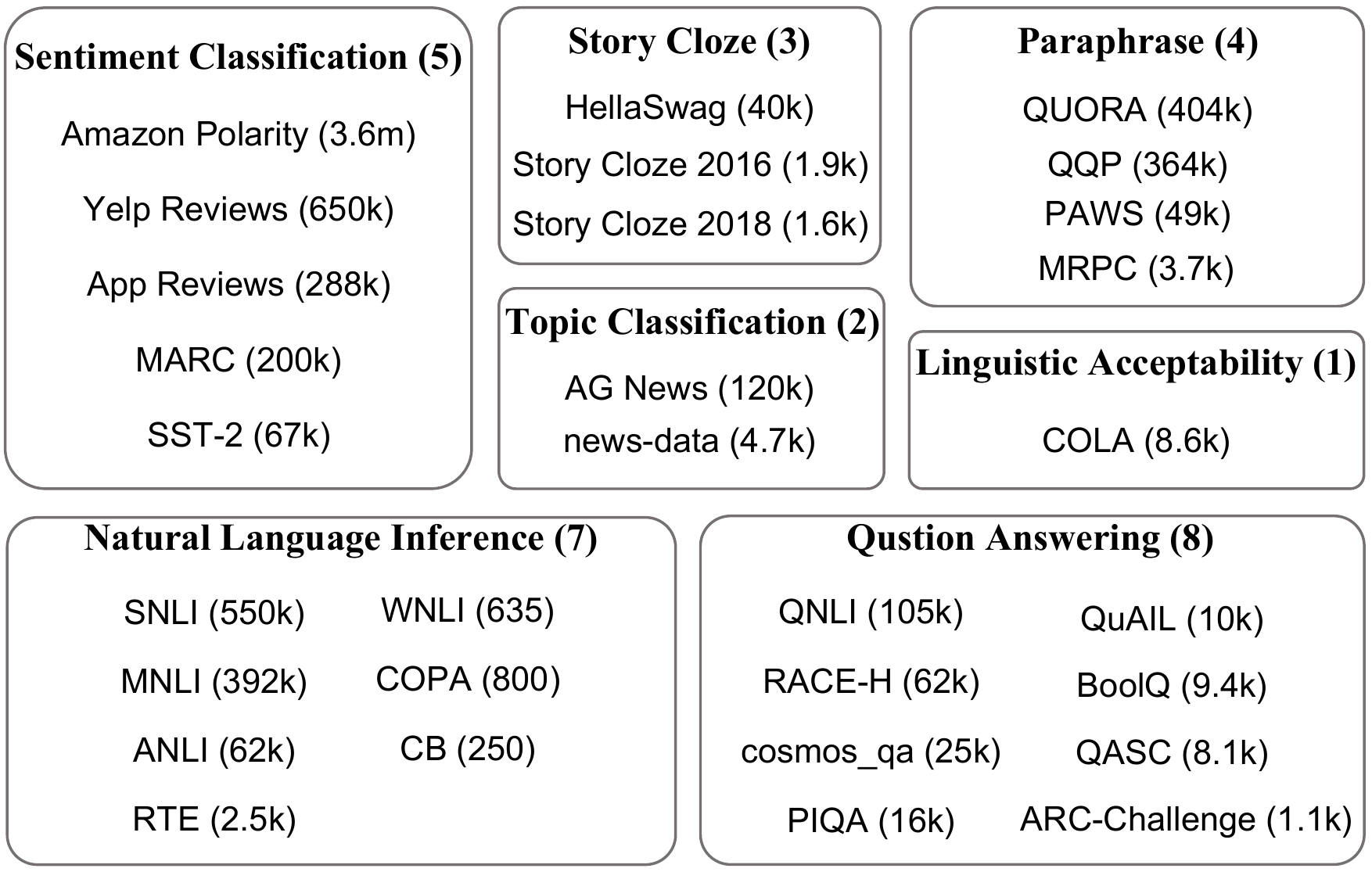}
    \caption{The list of \datasetnum datasets for 7 tasks; numbers in parentheses indicate dataset size.}
    \label{fig:datasets}
\end{figure}

We take the \textit{topic classification} task as an example to explain the mapping, also illustrated in  Figure~\ref{fig:template}. We convert the class label into a natural language description based on a template selected from a pre-designed template pool. In Figure~\ref{fig:template}, Template 1 is chosen among the $N$ templates.  Then we could build a candidate proposition statement by appending the natural language description to the original input text. The natural language description is the class labels in this example task. 

In our implementation, we utilize \datasetnum datasets of the seven chosen tasks (see Figure~\ref{fig:datasets}). For each dataset, we design at most $10$ templates to build the proposition.
Each item will randomly choose one template when building unified data. 
According to the chosen template, the original input context is concatenated with the label description to form a proposition. With the unified formatted proposition as input, our \modelname is responsible for giving a true-or-false judgment.

\modelname designs task unification by unified proposition format, which brings a good generalization effect. In other words, \modelname only has one proposition correctness judgment task (except language task), which endows \modelname with the ability to handle almost all NLU tasks.
As a comparison, multi-task needs special design for each task to fit the multi-task model. More importantly, multi-task could hardly gain generalization ability except the designed tasks.

\section{\modelname}
\label{sec:method}


\subsection{Architecture and Iterative Training}

\begin{figure}
    \centering
    \includegraphics[scale=0.85]{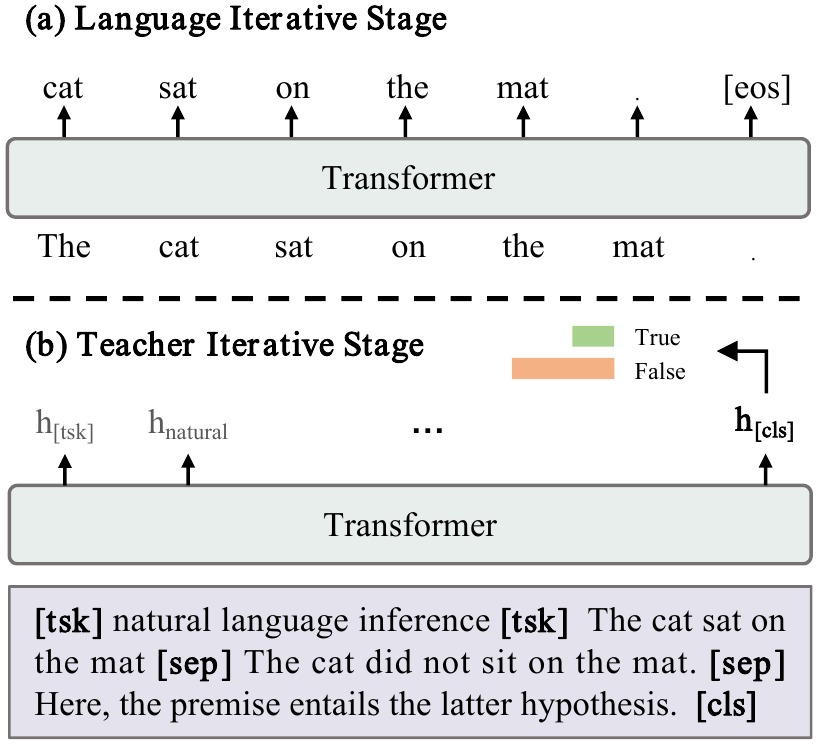}
    \caption{(a) The language stage utilizes the standard language training method, and (b) the teacher stage uses the unified proposition signal.}
    \label{fig:archi}
\end{figure}

Figure~\ref{fig:archi} shows the architecture of the proposed \modelname. The upper part of the figure shows training with language data, similar to the mainstream generative language models. The lower part shows training with task-aware unified data, through making correct  judgments on unified propositions. 

To train the model with both language and unified data, a straightforward choice is to first train the model fully on language data, then fully on unified data, similar to pre-train then fine-tuning. However, this strategy would lead the model to focus too much on tasks, and reduce its language capability and generalization. In our design, \textit{language signal} and \textit{teacher signal} are used for training in an iterative manner, analogous to  ``first read, then exercise''. 
Next, we detail the two training stages, which take turns to train the model.

\paragraph{Language Iterative Stage.} To keep the generation ability of language models, we choose the auto-regressive language model, more specifically GPT-2, as our base model in \modelname. 
As shown in Figure \ref{fig:archi}(a), \textit{language signal} guides the model to predict the next token with the previous token(s).
Given a text consists of $N$ tokens $\mathcal{T}=\{t_1, \ldots, t_N\}$, \modelname computes the probability $P_L$ of token $t_k$ with the previous $k-1$ tokens:
\begin{equation}
    P_L(t_k | t_1, \dots, t_{k-1}) = \softmax(W_vh_{t_{k-1}})
\end{equation}
Here, $h_{t_{k-1}}$ denotes the representation encoded by Transformer with the previous tokens $\{t_1, \ldots, t_{k-1}\}$ as input. $W_{v}$ represents the learnable parameters.

\paragraph{Teacher Iterative Stage.} In the proposed fine-tuning-free structure, \textit{teacher signal} aims to guide the model to learn task-oriented knowledge. After unifying all tasks to proposition correctness judgment, the model is expected to make correct judgements about the propositions. Let $u=(T_p,y)$ be a pair where $T_p$ refers to the proposition statement, and $y$ is the label of this proposition. When the proposition is right, $y$ is $\text{True}$, otherwise $\text{False}$.

Figure~\ref{fig:archi}(b) shows the mechanism of \textit{teacher signal} from unified data.
According to the designed transfer rule~(see Section~\ref{sec:unified}), each instance in the unified data is appended with a dedicated token ``[cls]''.
Given an instance, we can calculate the representation $h_{[\text{cls}]}$ of token ``[cls]'' by the shared Transformer of the language model. The proposition distribution $\mathcal{P}_P$ is computed by a $\text{softmax}$ classifier based on the learned representation $h_{[\text{cls}]}$:
\begin{equation}
    \mathcal{P}_P(y | T_p) = \softmax(W_ph_{[cls]}+b_p),
\end{equation}
where $W_p$ and $b_p$ are learnable parameters.

\subsection{Training Objective}

The training objective design has two parts. One is to maximize the likelihood of predicted tokens, defined in Equation~\ref{eqn:langObject}. The other is to minimize the cross-entropy of judging proposition correctness, defined in Equation~\ref{eqn:teacherObject}, where  $y$ is the ground truth of the given $T_p$.
\begin{align}
    J(\theta) &= \sum\sum_k \log P_L(t_k | t_1, \dots, t_{k-1}) \label{eqn:langObject} \\
    U(\theta) &= \sum \text{cross-entropy}\left(\mathcal{P}_P(y | T_p) , y\right)
    \label{eqn:teacherObject}
\end{align}

Note that the two objectives are alternately performed in each iteration, along with the two types of training data.

\section{Experiments}
\label{sec:main-exp}

\begin{table*}[htbp]
  \centering
    \resizebox{\textwidth}{!}{
    \begin{tabular}{l|ccccccccccc}
    \toprule
    \multicolumn{1}{c|}{\multirow{2}[2]{*}{Model}} & CoLA  & SST-2 & MRPC  & MRPC  & QQP   & QQP   & MNLI  & MNLI  & QNLI  & RTE   & Average \\
          & Mat. Corr & Acc.  & F1    & Acc.  & F1    & Acc.  & Mat. Acc. & M.m. Acc. & Acc.  & Acc.  & -- \\
    \midrule
    \multicolumn{12}{c}{\textit{Fine-Tuning}} \\
    \midrule
    \textcolor[rgb]{ .502,  .502,  .502}{GPT-2 \textit{w. F.T.}} & \textcolor[rgb]{ .502,  .502,  .502}{56.52} & \textcolor[rgb]{ .502,  .502,  .502}{94.95} & \textcolor[rgb]{ .502,  .502,  .502}{87.93} & \textcolor[rgb]{ .502,  .502,  .502}{82.84} & \textcolor[rgb]{ .502,  .502,  .502}{87.91} & \textcolor[rgb]{ .502,  .502,  .502}{90.83} & \textcolor[rgb]{ .502,  .502,  .502}{85.87} & \textcolor[rgb]{ .502,  .502,  .502}{85.83} & \textcolor[rgb]{ .502,  .502,  .502}{91.03} & \textcolor[rgb]{ .502,  .502,  .502}{68.23} & \textcolor[rgb]{ .502,  .502,  .502}{83.19} \\
    \midrule
    \multicolumn{12}{c}{\textit{Few-Shot \& Zero-Shot}} \\
    \midrule
    GPT-2 \textit{w. Z.S.} & -3.26 & 51.83 & 81.22 & 68.38 & 52.88 & 36.19 & 35.60  & 34.30  & 46.40  & 51.62 & 45.52 \\
    GPT-2 \textit{w. F.S.} & 0.60   & 63.30  & 81.34 & 68.62 & 44.46 & 46.80  & 31.60  & 35.00    & 48.60  & 49.09 & 46.94 \\
    GPT-3 \textit{w. Z.S.} & 17.28 & 52.98 & 65.05 & 53.92 & 4.61  & 62.80  & 36.80  & 39.50  & 60.90  & 64.98 & 45.88 \\
    GPT-3 \textit{w. F.S.} & -4.45 & 93.34 & 70.92 & 59.80  & 6.31  & 64.40  & 44.10  & 46.60  & 53.80  & 65.70  & 50.05 \\
    InstructGPT \textit{w. Z.S.} & \textbf{63.81} & 92.08 & 83.75 & 77.94 & 64.20  & 78.70  & 46.20  & 48.90  & 71.89 & \textbf{86.64} & 71.41 \\
    InstructGPT \textit{w. F.S.} & 56.72 & \textbf{95.18} & 77.41 & 70.83 & 72.98 & 81.20  & 71.70  & 72.89 & 81.30  & 82.67 & 76.29 \\
    \midrule
    \modelname & 53.00    & 93.81 & \textbf{87.21} & \textbf{82.11} & \textbf{85.47} & \textbf{89.60}  & \textbf{81.80}  & \textbf{79.80}  & \textbf{88.30}  & 71.12 & \textbf{81.22} \\
    \bottomrule
    \end{tabular}%
    }
      \caption{Comparison with GPT-2, GPT-3 and InstructGPT on GLUE validation set. \textit{F.T.}, \textit{Z.S.} and \textit{F.S.} denote fine-tuning, zero-shot and few-shot, repectively. Best results are in bold face (without considering GPT-2 w. F.T.).}

  \label{tab:main}%
\end{table*}%

We evaluate \modelname from two perspectives: language understanding performance, and language generation performance. Then we perform a detailed analysis of the impact of iterative processes, the impact of special tokens in the proposition format, and the generalization ability on unseen data.

\subsection{Training Datasets}
As indicated in Section~\ref{sec:unified}, for raw language data, we adopt the OpenWebText~\cite{Gokaslan2019OpenWeb}. The unified data are composed with \datasetnum datasets, listed in Figure~\ref{fig:datasets}, from the seven tasks.  Training sets are collected for all these datasets, except Story Cloze 2016 \& 2018, whose validation sets are collected.

For each instance in unified data, we construct a positive sample~(\ie true proposition) with the gold label according to the build method. Simultaneously, we randomly choose another label to construct a false proposition as a negative sample. 
In training set building, only one template would be randomly chosen for each original instance.
The maximum number of training set proposition samples per dataset is set to $100K$. The final unified training dataset contains about $1.76M$ samples.

\subsection{Evaluation on Language Understanding}
\label{sec:exp:understanding}

A key motivation of \modelname is the understanding ability without fine-tuning.
For understanding ability, we evaluate \modelname against strong baselines, including 175B GPT-3, InstructGPT,\footnote{We use text-davinci-003 version through OpenAI API.} and GPT-2, under the settings of zero-shot and few-shot on these baseline models. 

We mainly evaluate \modelname in \textit{fine-tuning-free} scenarios. That is, our \modelname does not conduct further task-oriented fine-tuning after training. Also, as an upper bound on the understanding ability with the same scale, we show the results of GPT-2 fine-tuning.

\paragraph{Evaluation Metric.}  We choose General Language Understanding Evaluation (GLUE) as the benchmark, which consists of typical natural language understanding tasks, including \textit{natural language inference}~(\ie AX, MNLI~\cite{DBLP:conf/naacl/WilliamsNB18}, RTE~\cite{DBLP:conf/tac/BentivogliMDDG09}, QNLI~\cite{DBLP:journals/corr/abs-1804-07461}, and WNLI~\cite{DBLP:conf/kr/LevesqueDM12}), \textit{sentiment classification}~(\ie SST-2~\cite{DBLP:conf/emnlp/SocherPWCMNP13} and STS-B~\cite{cer-etal-2017-semeval}), \textit{paraphrase}~(\ie MRPC~\cite{DBLP:conf/acl-iwp/DolanB05} and QQP) and \textit{linguistic acceptability}~(\ie CoLA).
Validation sets of each dataset are used for evaluation.
WNLI is excluded from the benchmark following the previous work~\cite{devlin-etal-2019-bert} because the GLUE webpage\footnote{\url{https://gluebenchmark.com/faq}} notes that there are issues with the construction of this dataset.
In addition, Ax is excluded due to the lack of the validation set. 
As the regression task (\ie STS-B) is not suitable for \modelname, we also exclude it from the benchmark.

We use the remaining seven tasks from the GLUE benchmark to evaluate. For experimental efficiency and cost effectiveness on API, when evaluating GPT-3 and InstructGPT, we use the validation set of GLUE for evaluation, and sample $1k$ instances for each task as the final evaluation set. All results are compared fairly in the same setting.

\paragraph{Results against Baselines.} The GLUE results are reported in Table~\ref{tab:main}. On both the average metric and 7 out of 10 evaluation metrics, \modelname outperforms all baselines, revealing its strong understanding ability. 
In particular, our \modelname gains about $5$ points improvement on average performance compared with the powerful InstructGPT. 
We also observe that MRPC, QQP, MNLI and QNLI tasks obtain $7$ or more points improvement in absolute accuracy. More importantly, the scale of our \modelname is only 0.3B. Even compared to the much larger GPT-3 with 175B parameters, \modelname achieves the state-of-the-art in the downstream tasks without fine-tuning, thanks to the strong task-aware signal in training. 
Among baseline models, InstructGPT has significant advantages over  GPT-2 and GPT-3. Our results also show that InstructGPT is an extremely strong model.

As reported in Table~\ref{tab:main}, our \modelname nearly matches the performance of \textit{GPT-2 with fine-tuning}. As GPT-2 is fine-tuned individually on each dataset, the comparison in essence is between one \modelname and $7$ dedicated GPT-2(s). In this sense, GPT-2 with fine-tuning could represent an approximate upper bound on 0.3B scale.
InstructGPT, the strongest generative PLM, achieves the second-best performance. As a comparison, \modelname uses classification-based approach while instruction tuning uses generation-based approach, which results in our model being more effective. More importantly, the task unification (see Section~\ref{sec:unified}) by one unified proposition format brings a good generalization effect.
An interesting point is that \modelname performs significantly worse than InstructGPT, but significantly better than GPT-2 with fine-tuning, on the accuracy of RTE. This result also suggests that the ability of language models could be positively related to the scale.

Not directly reflected in Table~\ref{tab:main}, through experiments, we also notice that \modelname is robust and insensitive for inference. 
Interestingly, we find that the results of the GPT family models predicted by few-shot or zero-shot are sensitive to the settings of templates, parameters, and post-processing methods. For example, we measure GPT-2 on the MRPC dataset with  a zero-shot setting. The results of the vocabulary probability distribution $top\_k=50$ are used to predict the classification. If we calculate the average sum of the probability ratios of positive and negative words, the accuracy is $33.08$. However, if we get the word by the highest probability value, the accuracy is more than doubled to $68.13$. The gap is so large that downstream tasks are hard to use reliably. On the other hand, \modelname is robust without the need of expert knowledge or parameter tuning.

\subsection{Evaluation on Language Generation}
\label{sec:exp:generation}

As aforementioned, while \modelname gains better understanding ability, we hope that the model retains its language generation ability, similar to GPT-2, for learning from the same language data. 
Hence, we evaluate the generation performance of \modelname against GPT-2. The evaluation is conducted on three public datasets, \ie LAMBADA~\cite{paperno-etal-2016-lambada}, WikiText2, and PTB. We use perplexity (PPL) to evaluate the generation performance, following common practice.

\begin{table}
  \centering
  \small{
    \begin{tabular}{l|ccc}
    \toprule
    \multicolumn{1}{c|}{Model} & LAMBADA & WikiText2 & PTB \\
    \midrule
    \modelname & 38.44 & 33.66 & 28.47 \\
    GPT-2  & 35.31 & 30.23 & 26.62 \\
    \bottomrule
    \end{tabular}%
    }
      \caption{Perplexity on \modelname and GPT-2.}

  \label{tab:ppl}%
\end{table}%

\begin{table*}
  \centering
  
  \resizebox{\textwidth}{!}{
    \begin{tabular}{p{21.915em}|p{20.585em}}
    \toprule
    \multicolumn{2}{l}{\textit{Case A}} \\
    \multicolumn{2}{p{42.5em}}{\textbf{INPUT:} Rick grew up in a troubled household. He never found good support in family, and turned to gangs. \textbf{It wasn't long before Rick got shot in a robbery. The incident caused him to turn a new leaf.}} \\
   \midrule
    \textbf{GPT-2:} "\textcolor{red}{I was a gang member, but I didn't know it at the time}," Rick said. "I had no idea what was going on. I just wanted to get away from it. That's how I ended up on the streets. And that's why I'm here today." & \textbf{\modelname:} "\textcolor{blue}{When I got out of prison, I was a different man}," he says. "I was able to get my life back on track. I went back to school, got a job, started a family. And that's what I've been doing ever since." \\
    \midrule
    \multicolumn{2}{l}{\textit{Case B}} \\
    \multicolumn{2}{p{42.5em}}{\textbf{INPUT:} Tony was happy to be going on his first cruise. He arrived at the docks and was ready with all his luggage. When he walked around the corner he couldn't believe the size of it. \textbf{He boarded and the ship left the docks.}} \\
    \midrule
    \textbf{GPT-2:} When he got to his room he was surprised to see that the room was bigger than he thought it would be. There was a bed, a desk, and a TV. ... & \textbf{\modelname:} "\textcolor{blue}{This is the first time I've ever been on a cruise ship}," he said. "I've never seen anything like this before." \\
    \midrule
    \multicolumn{2}{l}{\textit{Case C}} \\
    \multicolumn{2}{p{42.5em}}{\textbf{INPUT:} Jasper has been avoiding getting a real job for months. He has hated all the previous office jobs that he had. One day \textbf{he sees a job opening at his local butcher}. He applies successfully for the job at the butcher.} \\
    \midrule
    \textbf{GPT-2:} "He said, '\textcolor{red}{I don't want to hire you because you're too old},'" Jasper says. "And I was like, I'm not old. I've been doing this for a long time, and I know what it's like to be rejected. ... & \textbf{\modelname:} "\textcolor{blue}{It’s a great job},” he says. “I love it. I love the people here. They treat me like a king. The people are so nice. It makes me feel like I can do anything I want to do in this world. ... \\
    \bottomrule
    \end{tabular}%
    }
    \caption{Case study on generation ability. Consistency and inconsistency expressions are  in blue and red colors.}
  \label{tab:gen_case}
\end{table*}%

\paragraph{Results against Baselines.} 
Table~\ref{tab:ppl} reports the perplexity~(PPL) of \modelname and GPT-2. The PPL of \modelname is slightly higher than GPT-2. The minor gap could reflect that our model nearly reaches GPT-2 in terms of generation.  There are studies suggesting that PPL does not fully reflect the generation ability of language models~\cite{DBLP:journals/corr/abs-2210-05892}. Nevertheless, there are no good alternative metrics.

Since generation evaluation does not have a gold standard, 
we provide a case study on a story cloze task. 
The input texts come from the StoryCloze dataset~\cite{DBLP:journals/corr/MostafazadehCHP16}. In order to make the comparison objective and fair, we train a new \modelname model by removing this dataset from the unified data. $\text{Top-k}$ random sampling with $k = 50$ is used for generation, width of $5$ for beam search, and the maximum generated length is $256$. 
Table~\ref{tab:gen_case} shows three generated samples.

First, we find that the readability of the text generated by \modelname is good by manual sampling inspection. There are no obvious grammatical errors, which shows that \modelname and GPT-2 perform at the same level in the grammatical aspect.
Second, Table~\ref{tab:gen_case} shows that there are many context inconsistency phenomena in the texts generated by GPT-2. Our \modelname could generate more consistent text with the same texts as input. 
To clearly show the correct logical association, we use the text in blue to highlight the relation. Meanwhile, we use the text in red to warn the logical errors. 
Cases A and B show that our \modelname is capable of generating more consistent texts. Interestingly, although there is no obvious grammatical error in Case B, the generated text by GPT-2 almost has no direct logical relationship with the input text. That is, it is not a high-quality generated result.

In simple words, our experiments show that the proposed \modelname achieves the same level of PPL as GPT-2. \modelname could generate more consistent text. We believe that if we train our model at a larger scale, better generation ability can be achieved.

\subsection{Detailed Analysis}

We conduct three sets of experiments to study (i) the impact of the iterative training, (ii) the proposition format, and (iii) the generalization ability on unseen tasks.

\begin{table*}
  \centering
 
  \resizebox{\textwidth}{!}{
    \begin{tabular}{l|ccccccccccc}
    \toprule
    \multicolumn{1}{c|}{\multirow{2}[2]{*}{Model}} & CoLA  & SST-2 & MRPC  & MRPC  & QQP   & QQP   & MNLI  & MNLI  & QNLI  & RTE & Average\\
          & Mat. Corr & Acc.  & F1    & Acc.  & F1    & Acc.  & Mat. Acc & M.m. Acc & Acc.  & Acc. & --\\
    \midrule
    \modelname & \textbf{52.53} & \textbf{93.81} & \textbf{87.21} & \textbf{82.11} & \textbf{86.32} & \textbf{89.82} & \textbf{80.70}  & \textbf{80.93} & \textbf{88.61} & 71.12 & \textbf{81.32}\\
    \ \ \textit{w.o. L.S.} & 40.14 & 92.43 & 85.71 & 79.17 & 85.66 & 89.11 & 77.56 & 78.57 & 87.28 & \textbf{71.48} & 78.71 \\
    \bottomrule
    \end{tabular}%
    }
     \caption{The effect of \textit{language signal} for understanding, where \textit{L.S.} refers to \textit{language signal}.}
  \label{tab:without-language}%
  \vspace{1ex}
\end{table*}

\begin{table*}[htbp]
  \centering
  
  \resizebox{\textwidth}{!}{
    \begin{tabular}{l|ccccccccccc}
    \toprule
    \multicolumn{1}{c|}{\multirow{2}[2]{*}{Model}} & CoLA  & SST-2 & MRPC  & MRPC  & QQP   & QQP   & MNLI  & MNLI  & QNLI  & RTE & Average\\
          & Mat. Corr & Acc.  & F1    & Acc.  & F1    & Acc.  & Mat. Acc & M.m. Acc & Acc.  & Acc. & --\\
    \midrule
    \modelname & \textbf{52.53} & \textbf{93.81} & \textbf{87.21} & \textbf{82.11} & \textbf{86.32} & \textbf{89.82} & \textbf{80.70}  & \textbf{80.93} & 88.61 & 71.12 & \textbf{81.32}\\
    \ \ \textit{w.o. prefix} & 48.15 & 92.78 & 86.76 & 81.37 & 86.06 & 89.53 & 80.23 & 80.32 & \textbf{88.76} & \textbf{71.84} & 80.58\\
    \bottomrule
    \end{tabular}%
}
\caption{The effect of task prefix for understanding.}
  \label{tab:without-prefix}%
  \vspace{1ex}
\end{table*}%

\begin{table}
  \centering
  \small
  
  \scriptsize\setlength{\tabcolsep}{1.7mm}{
    \begin{tabular}{l|cccccc}
    \toprule
    \multicolumn{1}{c|}{\multirow{2}[2]{*}{Model}} & MRPC  & MRPC  & RTE   & COPA  & CB    & Average \\
          & F1    & Acc.  & Acc.  & Acc.  & Acc.  & -- \\
    \midrule
    
    \modelname & 87.21 & 82.11 & 71.12 & 73.00    & 73.21 & 77.33 \\
    $\text{\modelname}_\text{U}$ & 81.22 & 72.06 & 66.79 & 68.00    & 68.64 & 71.34 \\
    GPT-2 \textit{w. F.T.} & 87.93 & 82.84 & 68.23 & 54.00    & 73.21 & 73.24 \\
    \bottomrule
    \end{tabular}%
    }
    \caption{\modelname vs. GPT-2 fine-tuning on unseen data.}
  \label{tab:unseen_eval}%
\end{table}%

\paragraph{The Iterative Training.} 
If we remove the \textit{teacher signal}, \modelname will degenerate into a general language model. Based on the results in Section~\ref{sec:exp:understanding} and \ref{sec:exp:generation}, we conclude that the discarding of the \textit{teacher signal} will lead to a reduction in the performance of understanding tasks.
At the opposite extreme, we remove the \textit{language signal}. The model could then only rely on the objective of proposition correctness judgment for training. The comparison is under the same training parameters and the same number of steps. 
Table~\ref{tab:without-language} shows the results compared to standard \modelname. After removing \textit{language signal}, almost all evaluation metrics show a significant drop. Interestingly, the PPL of WikiText2 could increase to $8,000$, even $9,000$, compared to a regular scale $30$ from GPT-2. As a result, the generation ability is damaged seriously.

\begin{figure}
    \centering
    \includegraphics[scale=0.5]{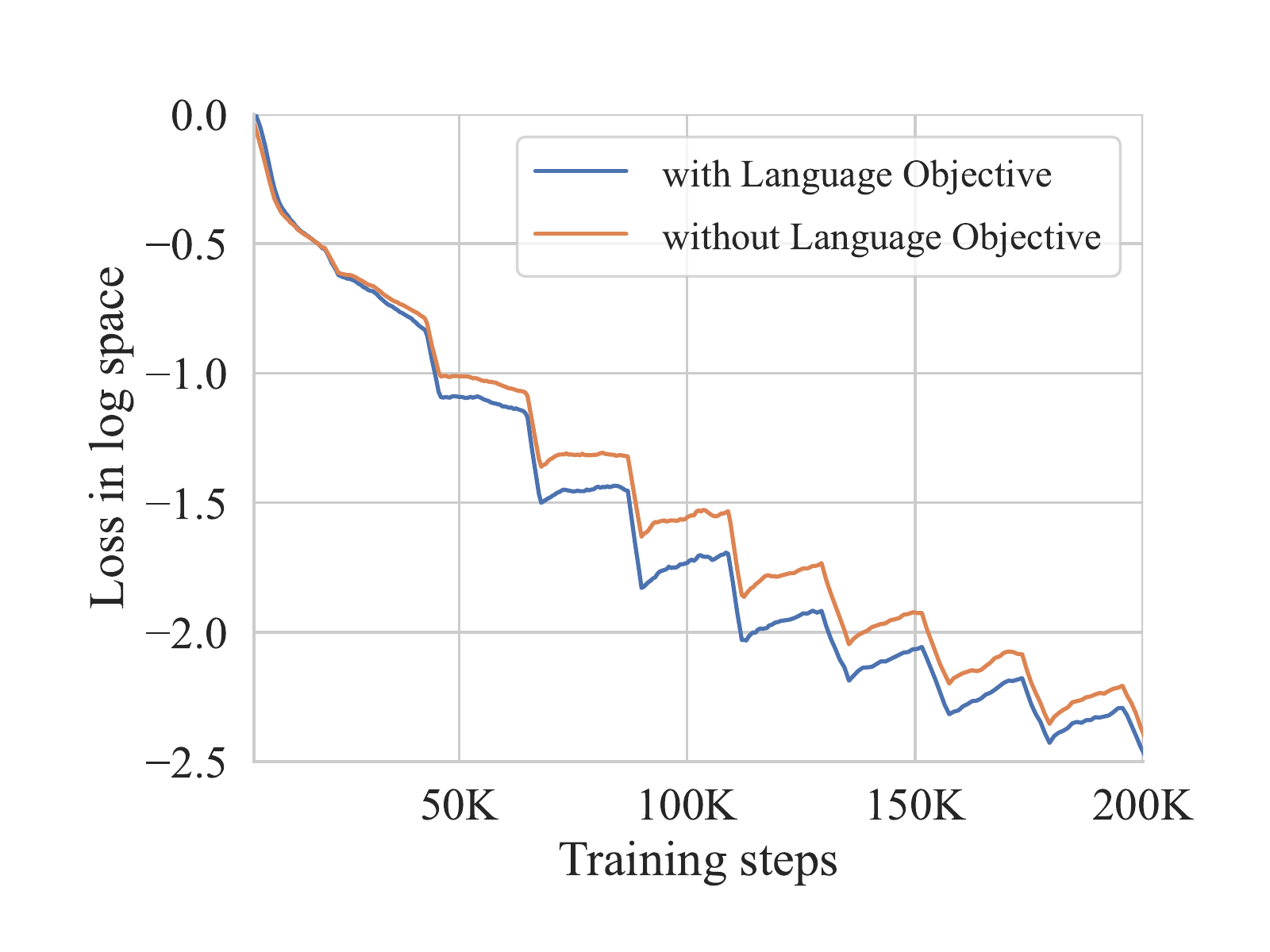}
    \caption{Effect of \textit{language signal} on proposition loss.}
    \label{fig:mtloss}
\end{figure}

We have shown that \textit{language signal} could help \modelname learn better. Also, to evaluate whether \modelname can learn faster, we give the loss performance during training in Figure~\ref{fig:mtloss}. We observe that after $20k$ steps, the proposition loss of \modelname decreases faster. It shows that the \textit{language signal} has a positive effect on task-oriented learning. 
In conclusion, \textit{language signal} and \textit{teacher signal} could promote each other under the iterative training method.

\paragraph{The Proposition Format.} 
Task prefix, such as ``[tsk] Topic Classification [tsk]'', could guide the model to narrow down the search space. 
In the implementation, the set of task prefix is \{\textit{Linguistic Acceptability}, \textit{Topic Classification}, \textit{Story Cloze}, \textit{Sentiment Classification}, \textit{Question Answering}, \textit{Paraphrase}, \textit{Natural Language Inference}\}.
Since the parameters of \modelname are small, we utilize this kind of task prefix to guide the model when meeting task-oriented understanding. To compare, we train another \modelname without task prefix. Other settings are the same as \modelname. Table~\ref{tab:without-prefix} gives the comparison results.

We observe that the performance of most tasks is at a similar level. This phenomenon could prove that \modelname does not significantly  rely on task prefix. We believe that when the size of \modelname exceeds $1B$, task prefix will have little effect.


\paragraph{Generalization to Unseen Dataset.} 
To evaluate the generalization ability of \modelname, we train a new model $\text{\modelname}_\text{U}$ by removing 4 datasets from unified data. They are MRPC, RTE, COPA~\cite{gordon-etal-2012-semeval} and CB~\cite{de2019commitmentbank}. Other settings are exactly the same. Table~\ref{tab:unseen_eval} gives the results compared to standard \modelname.

We observe that the average score of $\text{\modelname}_\text{U}$ nearly matches the performance of \textit{GPT-2 with fine-tuning}. 
More interestingly, the accuracy of the $\text{\modelname}_\text{U}$ on  the COPA dataset has increased by $14$ points than \textit{GPT-2 with fine-tuning}. Compared to the standard \modelname, the average drop is about $6$ points. It is an inspiring result because it reveals that our \modelname has a good generalization ability.
We believe the reason is that we unify various tasks into one proposition correctness judgment task. The phenomenon also reminds us that the design of \textit{teacher signal} is a key to training a better language model.

\section{Conclusion}
\label{sec:conclusion}

With the aim of reducing costs in training and deployment, we design a novel fine-tuning-free language model. The model training benefits from the self-supervised \textit{language signal} as a typical language model does. It also  becomes task-aware through the training on unified data. Evaluations  show that \modelname retains its strong language capability as a language model in a similar scale \ie GPT-2.   Experiments also show that \modelname's  understanding performance on the pre-defined tasks is significantly better than GPT-3 and instructGPT with zero- or few-shot settings, even though our model is much smaller,  0.3B versus 175B parameters. \modelname also generalizes well if a task can also be mapped to the unified format. 
We believe the key reason behind the strong performance of \modelname is the task unification and the alternative training. If an enterprise can design a similar unified data format for all their tasks, like the proposition format in our setting, the proposed fine-tuning-free training strategy could be a good consideration for cost saving.

\section*{Limitation}

There are three main limitations to this work. First, it is interesting to study whether our model could lower the data size for training. If this interesting point could be supported,  computing costs could be further reduced, and the large model could be greener. 
Another interesting point is to study the impact of the order of task data. Lastly, we believe that our proposed model could work better on a larger scale, but we do not prove that due to the extremely high costs. 

\section*{Acknowledgments}

This work was supported by the National Key R\&D Program of China (2022ZD0116300) and the National Science Foundation of China (NSFC No. 62106249).

\bibliography{anthology,custom}

\begin{thebibliography}{30}
\expandafter\ifx\csname natexlab\endcsname\relax\def\natexlab#1{#1}\fi

\bibitem[{Bentivogli et~al.(2009)Bentivogli, Magnini, Dagan, Dang, and
  Giampiccolo}]{DBLP:conf/tac/BentivogliMDDG09}
Luisa Bentivogli, Bernardo Magnini, Ido Dagan, Hoa~Trang Dang, and Danilo
  Giampiccolo. 2009.
\newblock \href
  {https://tac.nist.gov/publications/2009/additional.papers/RTE5\_overview.proceedings.pdf}
  {The fifth {PASCAL} recognizing textual entailment challenge}.
\newblock In \emph{Proceedings of the Second Text Analysis Conference, {TAC}
  2009, Gaithersburg, Maryland, USA, November 16-17, 2009}. {NIST}.

\bibitem[{Brown et~al.(2020)Brown, Mann, Ryder, Subbiah, Kaplan, Dhariwal,
  Neelakantan, Shyam, Sastry, Askell et~al.}]{brown2020language}
Tom Brown, Benjamin Mann, Nick Ryder, Melanie Subbiah, Jared~D Kaplan, Prafulla
  Dhariwal, Arvind Neelakantan, Pranav Shyam, Girish Sastry, Amanda Askell,
  et~al. 2020.
\newblock Language models are few-shot learners.
\newblock \emph{Advances in neural information processing systems},
  33:1877--1901.

\bibitem[{Cer et~al.(2017)Cer, Diab, Agirre, Lopez-Gazpio, and
  Specia}]{cer-etal-2017-semeval}
Daniel Cer, Mona Diab, Eneko Agirre, I{\~n}igo Lopez-Gazpio, and Lucia Specia.
  2017.
\newblock \href {https://doi.org/10.18653/v1/S17-2001} {{S}em{E}val-2017 task
  1: Semantic textual similarity multilingual and crosslingual focused
  evaluation}.
\newblock In \emph{Proceedings of the 11th International Workshop on Semantic
  Evaluation ({S}em{E}val-2017)}, pages 1--14, Vancouver, Canada. Association
  for Computational Linguistics.

\bibitem[{Dai et~al.(2019)Dai, Yang, Yang, Carbonell, Le, and
  Salakhutdinov}]{DBLP:conf/acl/DaiYYCLS19}
Zihang Dai, Zhilin Yang, Yiming Yang, Jaime~G. Carbonell, Quoc~Viet Le, and
  Ruslan Salakhutdinov. 2019.
\newblock \href {https://doi.org/10.18653/v1/p19-1285} {Transformer-xl:
  Attentive language models beyond a fixed-length context}.
\newblock In \emph{Proceedings of the 57th Conference of the Association for
  Computational Linguistics, {ACL} 2019, Florence, Italy, July 28- August 2,
  2019, Volume 1: Long Papers}, pages 2978--2988. Association for Computational
  Linguistics.

\bibitem[{De~Marneffe et~al.(2019)De~Marneffe, Simons, and
  Tonhauser}]{de2019commitmentbank}
Marie-Catherine De~Marneffe, Mandy Simons, and Judith Tonhauser. 2019.
\newblock The commitmentbank: Investigating projection in naturally occurring
  discourse.
\newblock In \emph{proceedings of Sinn und Bedeutung}, volume~23, pages
  107--124.

\bibitem[{Devlin et~al.(2019)Devlin, Chang, Lee, and
  Toutanova}]{devlin-etal-2019-bert}
Jacob Devlin, Ming-Wei Chang, Kenton Lee, and Kristina Toutanova. 2019.
\newblock \href {https://doi.org/10.18653/v1/N19-1423} {{BERT}: Pre-training of
  deep bidirectional transformers for language understanding}.
\newblock In \emph{Proceedings of the 2019 Conference of the North {A}merican
  Chapter of the Association for Computational Linguistics: Human Language
  Technologies, Volume 1 (Long and Short Papers)}, pages 4171--4186,
  Minneapolis, Minnesota. Association for Computational Linguistics.

\bibitem[{Dolan and Brockett(2005)}]{DBLP:conf/acl-iwp/DolanB05}
William~B. Dolan and Chris Brockett. 2005.
\newblock \href {https://aclanthology.org/I05-5002/} {Automatically
  constructing a corpus of sentential paraphrases}.
\newblock In \emph{Proceedings of the Third International Workshop on
  Paraphrasing, IWP@IJCNLP 2005, Jeju Island, Korea, October 2005, 2005}. Asian
  Federation of Natural Language Processing.

\bibitem[{Du et~al.(2022)Du, Qian, Liu, Ding, Qiu, Yang, and Tang}]{GLM}
Zhengxiao Du, Yujie Qian, Xiao Liu, Ming Ding, Jiezhong Qiu, Zhilin Yang, and
  Jie Tang. 2022.
\newblock \href {https://doi.org/10.18653/v1/2022.acl-long.26} {{GLM:} general
  language model pretraining with autoregressive blank infilling}.
\newblock In \emph{Proceedings of the 60th Annual Meeting of the Association
  for Computational Linguistics (Volume 1: Long Papers), {ACL} 2022, Dublin,
  Ireland, May 22-27, 2022}, pages 320--335. Association for Computational
  Linguistics.

\bibitem[{Gokaslan and Cohen(2019)}]{Gokaslan2019OpenWeb}
Aaron Gokaslan and Vanya Cohen. 2019.
\newblock Openwebtext corpus.
\newblock \url{http://Skylion007.github.io/OpenWebTextCorpus}.

\bibitem[{Gordon et~al.(2012)Gordon, Kozareva, and
  Roemmele}]{gordon-etal-2012-semeval}
Andrew Gordon, Zornitsa Kozareva, and Melissa Roemmele. 2012.
\newblock \href {https://aclanthology.org/S12-1052} {{S}em{E}val-2012 task 7:
  Choice of plausible alternatives: An evaluation of commonsense causal
  reasoning}.
\newblock In \emph{*{SEM} 2012: The First Joint Conference on Lexical and
  Computational Semantics {--} Volume 1: Proceedings of the main conference and
  the shared task, and Volume 2: Proceedings of the Sixth International
  Workshop on Semantic Evaluation ({S}em{E}val 2012)}, pages 394--398,
  Montr{\'e}al, Canada. Association for Computational Linguistics.

\bibitem[{Grinsztajn et~al.(2022)Grinsztajn, Oyallon, and
  Varoquaux}]{DBLP:journals/corr/abs-2207-08815}
L{\'{e}}o Grinsztajn, Edouard Oyallon, and Ga{\"{e}}l Varoquaux. 2022.
\newblock \href {https://doi.org/10.48550/arXiv.2207.08815} {Why do tree-based
  models still outperform deep learning on tabular data?}
\newblock \emph{CoRR}, abs/2207.08815.

\bibitem[{Joshi et~al.(2020)Joshi, Chen, Liu, Weld, Zettlemoyer, and
  Levy}]{DBLP:journals/tacl/JoshiCLWZL20}
Mandar Joshi, Danqi Chen, Yinhan Liu, Daniel~S. Weld, Luke Zettlemoyer, and
  Omer Levy. 2020.
\newblock \href {https://doi.org/10.1162/tacl\_a\_00300} {Spanbert: Improving
  pre-training by representing and predicting spans}.
\newblock \emph{Trans. Assoc. Comput. Linguistics}, 8:64--77.

\bibitem[{Levesque et~al.(2012)Levesque, Davis, and
  Morgenstern}]{DBLP:conf/kr/LevesqueDM12}
Hector~J. Levesque, Ernest Davis, and Leora Morgenstern. 2012.
\newblock \href {http://www.aaai.org/ocs/index.php/KR/KR12/paper/view/4492}
  {The winograd schema challenge}.
\newblock In \emph{Principles of Knowledge Representation and Reasoning:
  Proceedings of the Thirteenth International Conference, {KR} 2012, Rome,
  Italy, June 10-14, 2012}. {AAAI} Press.

\bibitem[{Liu et~al.(2019)Liu, Ott, Goyal, Du, Joshi, Chen, Levy, Lewis,
  Zettlemoyer, and Stoyanov}]{DBLP:journals/corr/abs-1907-11692}
Yinhan Liu, Myle Ott, Naman Goyal, Jingfei Du, Mandar Joshi, Danqi Chen, Omer
  Levy, Mike Lewis, Luke Zettlemoyer, and Veselin Stoyanov. 2019.
\newblock \href {http://arxiv.org/abs/1907.11692} {Roberta: {A} robustly
  optimized {BERT} pretraining approach}.
\newblock \emph{CoRR}, abs/1907.11692.

\bibitem[{Mostafazadeh et~al.(2016)Mostafazadeh, Chambers, He, Parikh, Batra,
  Vanderwende, Kohli, and Allen}]{DBLP:journals/corr/MostafazadehCHP16}
Nasrin Mostafazadeh, Nathanael Chambers, Xiaodong He, Devi Parikh, Dhruv Batra,
  Lucy Vanderwende, Pushmeet Kohli, and James~F. Allen. 2016.
\newblock \href {http://arxiv.org/abs/1604.01696} {A corpus and evaluation
  framework for deeper understanding of commonsense stories}.
\newblock \emph{CoRR}, abs/1604.01696.

\bibitem[{Nie et~al.(2021)Nie, Miao, Cao, Ma, Liu, Xue, Miao, Liu, Yang, and
  Cui}]{MOE2}
Xiaonan Nie, Xupeng Miao, Shijie Cao, Lingxiao Ma, Qibin Liu, Jilong Xue,
  Youshan Miao, Yi~Liu, Zhi Yang, and Bin Cui. 2021.
\newblock \href {https://doi.org/10.48550/ARXIV.2112.14397} {Evomoe: An
  evolutional mixture-of-experts training framework via dense-to-sparse gate}.

\bibitem[{Ouyang et~al.(2022)Ouyang, Wu, Jiang, Almeida, Wainwright, Mishkin,
  Zhang, Agarwal, Slama, Ray, Schulman, Hilton, Kelton, Miller, Simens, Askell,
  Welinder, Christiano, Leike, and
  Lowe}]{https://doi.org/10.48550/arxiv.2203.02155}
Long Ouyang, Jeff Wu, Xu~Jiang, Diogo Almeida, Carroll~L. Wainwright, Pamela
  Mishkin, Chong Zhang, Sandhini Agarwal, Katarina Slama, Alex Ray, John
  Schulman, Jacob Hilton, Fraser Kelton, Luke Miller, Maddie Simens, Amanda
  Askell, Peter Welinder, Paul Christiano, Jan Leike, and Ryan Lowe. 2022.
\newblock \href {https://doi.org/10.48550/ARXIV.2203.02155} {Training language
  models to follow instructions with human feedback}.

\bibitem[{Paperno et~al.(2016)Paperno, Kruszewski, Lazaridou, Pham, Bernardi,
  Pezzelle, Baroni, Boleda, and Fern{\'a}ndez}]{paperno-etal-2016-lambada}
Denis Paperno, Germ{\'a}n Kruszewski, Angeliki Lazaridou, Ngoc~Quan Pham,
  Raffaella Bernardi, Sandro Pezzelle, Marco Baroni, Gemma Boleda, and Raquel
  Fern{\'a}ndez. 2016.
\newblock \href {https://doi.org/10.18653/v1/P16-1144} {The {LAMBADA} dataset:
  Word prediction requiring a broad discourse context}.
\newblock In \emph{Proceedings of the 54th Annual Meeting of the Association
  for Computational Linguistics (Volume 1: Long Papers)}, pages 1525--1534,
  Berlin, Germany. Association for Computational Linguistics.

\bibitem[{Radford et~al.(2018)Radford, Narasimhan, Salimans, Sutskever
  et~al.}]{radford2018improving}
Alec Radford, Karthik Narasimhan, Tim Salimans, Ilya Sutskever, et~al. 2018.
\newblock Improving language understanding by generative pre-training.

\bibitem[{Radford et~al.(2019)Radford, Wu, Child, Luan, Amodei, and
  Sutskever}]{radford2019language}
Alec Radford, Jeff Wu, Rewon Child, David Luan, Dario Amodei, and Ilya
  Sutskever. 2019.
\newblock Language models are unsupervised multitask learners.

\bibitem[{Raffel et~al.(2020)Raffel, Shazeer, Roberts, Lee, Narang, Matena,
  Zhou, Li, and Liu}]{DBLP:journals/jmlr/RaffelSRLNMZLL20}
Colin Raffel, Noam Shazeer, Adam Roberts, Katherine Lee, Sharan Narang, Michael
  Matena, Yanqi Zhou, Wei Li, and Peter~J. Liu. 2020.
\newblock \href {http://jmlr.org/papers/v21/20-074.html} {Exploring the limits
  of transfer learning with a unified text-to-text transformer}.
\newblock \emph{J. Mach. Learn. Res.}, 21:140:1--140:67.

\bibitem[{Sanh et~al.(2019)Sanh, Debut, Chaumond, and Wolf}]{distilbert}
Victor Sanh, Lysandre Debut, Julien Chaumond, and Thomas Wolf. 2019.
\newblock \href {http://arxiv.org/abs/1910.01108} {Distilbert, a distilled
  version of {BERT:} smaller, faster, cheaper and lighter}.
\newblock \emph{CoRR}, abs/1910.01108.

\bibitem[{Socher et~al.(2013)Socher, Perelygin, Wu, Chuang, Manning, Ng, and
  Potts}]{DBLP:conf/emnlp/SocherPWCMNP13}
Richard Socher, Alex Perelygin, Jean Wu, Jason Chuang, Christopher~D. Manning,
  Andrew~Y. Ng, and Christopher Potts. 2013.
\newblock \href {https://aclanthology.org/D13-1170/} {Recursive deep models for
  semantic compositionality over a sentiment treebank}.
\newblock In \emph{Proceedings of the 2013 Conference on Empirical Methods in
  Natural Language Processing, {EMNLP} 2013, 18-21 October 2013, Grand Hyatt
  Seattle, Seattle, Washington, USA, {A} meeting of SIGDAT, a Special Interest
  Group of the {ACL}}, pages 1631--1642. {ACL}.

\bibitem[{Sun et~al.(2019)Sun, Wang, Li, Feng, Chen, Zhang, Tian, Zhu, Tian,
  and Wu}]{DBLP:journals/corr/abs-1904-09223}
Yu~Sun, Shuohuan Wang, Yu{-}Kun Li, Shikun Feng, Xuyi Chen, Han Zhang, Xin
  Tian, Danxiang Zhu, Hao Tian, and Hua Wu. 2019.
\newblock \href {http://arxiv.org/abs/1904.09223} {{ERNIE:} enhanced
  representation through knowledge integration}.
\newblock \emph{CoRR}, abs/1904.09223.

\bibitem[{Wang et~al.(2018)Wang, Singh, Michael, Hill, Levy, and
  Bowman}]{DBLP:journals/corr/abs-1804-07461}
Alex Wang, Amanpreet Singh, Julian Michael, Felix Hill, Omer Levy, and
  Samuel~R. Bowman. 2018.
\newblock \href {http://arxiv.org/abs/1804.07461} {{GLUE:} {A} multi-task
  benchmark and analysis platform for natural language understanding}.
\newblock \emph{CoRR}, abs/1804.07461.

\bibitem[{Wang et~al.(2021)Wang, Fang, Khabsa, Mao, and
  Ma}]{https://doi.org/10.48550/arxiv.2104.14690}
Sinong Wang, Han Fang, Madian Khabsa, Hanzi Mao, and Hao Ma. 2021.
\newblock \href {https://doi.org/10.48550/ARXIV.2104.14690} {Entailment as
  few-shot learner}.

\bibitem[{Wang et~al.(2022)Wang, Deng, Sun, and
  Meng}]{DBLP:journals/corr/abs-2210-05892}
Yequan Wang, Jiawen Deng, Aixin Sun, and Xuying Meng. 2022.
\newblock \href {https://doi.org/10.48550/arXiv.2210.05892} {Perplexity from
  {PLM} is unreliable for evaluating text quality}.
\newblock \emph{CoRR}, abs/2210.05892.

\bibitem[{Wei et~al.(2022)Wei, Bosma, Zhao, Guu, Yu, Lester, Du, Dai, and
  Le}]{DBLP:conf/iclr/WeiBZGYLDDL22}
Jason Wei, Maarten Bosma, Vincent~Y. Zhao, Kelvin Guu, Adams~Wei Yu, Brian
  Lester, Nan Du, Andrew~M. Dai, and Quoc~V. Le. 2022.
\newblock \href {https://openreview.net/forum?id=gEZrGCozdqR} {Finetuned
  language models are zero-shot learners}.
\newblock In \emph{The Tenth International Conference on Learning
  Representations, {ICLR} 2022, Virtual Event, April 25-29, 2022}.
  OpenReview.net.

\bibitem[{Williams et~al.(2018)Williams, Nangia, and
  Bowman}]{DBLP:conf/naacl/WilliamsNB18}
Adina Williams, Nikita Nangia, and Samuel~R. Bowman. 2018.
\newblock \href {https://doi.org/10.18653/v1/n18-1101} {A broad-coverage
  challenge corpus for sentence understanding through inference}.
\newblock In \emph{Proceedings of the 2018 Conference of the North American
  Chapter of the Association for Computational Linguistics: Human Language
  Technologies, {NAACL-HLT} 2018, New Orleans, Louisiana, USA, June 1-6, 2018,
  Volume 1 (Long Papers)}, pages 1112--1122. Association for Computational
  Linguistics.

\bibitem[{Yang et~al.(2019)Yang, Dai, Yang, Carbonell, Salakhutdinov, and
  Le}]{DBLP:conf/nips/YangDYCSL19}
Zhilin Yang, Zihang Dai, Yiming Yang, Jaime~G. Carbonell, Ruslan Salakhutdinov,
  and Quoc~V. Le. 2019.
\newblock \href
  {https://proceedings.neurips.cc/paper/2019/hash/dc6a7e655d7e5840e66733e9ee67cc69-Abstract.html}
  {Xlnet: Generalized autoregressive pretraining for language understanding}.
\newblock In \emph{Advances in Neural Information Processing Systems 32: Annual
  Conference on Neural Information Processing Systems 2019, NeurIPS 2019,
  December 8-14, 2019, Vancouver, BC, Canada}, pages 5754--5764.

\end{thebibliography}
\bibliographystyle{acl_natbib}

\appendix

\section{Appendix}
\label{sec:appendix}
\subsection{Implementation Detail}

We list the implementation details of the proposed \modelname to support reproduction. To support research, we will release the corresponding codes and trained parameters upon acceptance.

In our experiment of training \modelname, we choose the medium version of GPT-2 with 345M parameters, which makes up of 12 transformer blocks. The model dimension is 768. 
To speed up training and save energy, we initialize the model with the pre-trained parameters provided by HuggingFace.
All the experiments are performed on 5 nodes each with eight 40GB NVIDIA Tesla A100~(40 cards total) for about 8 days.
We train \modelname for 8 epochs on OpenWebText with a batch size of 4 on each card. Each epoch takes around 13.5 to 17.5 hours to complete. We use AdamW optimizer with a learning rate of 5e-5.
The mixing ratio of the language iterative stage and teacher iterative stage samples is 2:1.

\end{document}